\documentclass[11pt]{article}
\usepackage[T1]{fontenc}
\usepackage[utf8]{inputenc}
\usepackage{mathptmx}
\usepackage[scaled=0.92]{helvet}
\usepackage{courier}
\usepackage{geometry}
\usepackage{amsmath,amsfonts,amssymb}
\usepackage{booktabs}
\usepackage{graphicx}
\usepackage{placeins}
\usepackage{xcolor}
\usepackage[numbers,sort&compress]{natbib}
\usepackage{algorithm}
\usepackage{algorithmic}
\usepackage{multirow}
\usepackage{caption}
\usepackage{enumitem}
\usepackage{fancyhdr}
\usepackage{titlesec}
\usepackage{nicefrac}
\usepackage{url}
\usepackage[colorlinks=true,linkcolor=blue!70!black,citecolor=blue!70!black,urlcolor=blue!70!black]{hyperref}

\titleformat{\section}{\large\bfseries}{\thesection}{0.8em}{}
\titleformat{\subsection}{\normalsize\bfseries}{\thesubsection}{0.8em}{}
\titlespacing{\section}{0pt}{12pt plus 2pt minus 1pt}{6pt plus 1pt}
\titlespacing{\subsection}{0pt}{8pt plus 2pt minus 1pt}{4pt plus 1pt}

\newcommand{\method}{\textsc{LogicDiff}}
\newcommand{\premise}{\textsc{Premise}}
\newcommand{\connective}{\textsc{Connective}}
\newcommand{\derived}{\textsc{Derived}}
\newcommand{\conclusion}{\textsc{Conclusion}}
\newcommand{\filler}{\textsc{Filler}}
\newcommand{\pp}{\,\text{pp}}

\begin{document}

\begin{center}
{\LARGE\bfseries LogicDiff: Logic-Guided Denoising Improves Reasoning\\ in Masked Diffusion Language Models\par}
\vspace{14pt}
{\large Shaik Aman}\\[2pt]
{\normalsize Independent Researcher}\\
{\normalsize Nellore, Andhra Pradesh, India}\\
{\small\texttt{amanabdul21@gmail.com}}
\vspace{16pt}
\end{center}

\begin{center}\textbf{Abstract}\end{center}
\vspace{-4pt}
{\small
Masked diffusion language models (MDLMs) generate text by iteratively unmasking tokens from a fully masked sequence. Their standard confidence-based unmasking strategy systematically defers high-entropy logical connective tokens, degrading reasoning performance. We introduce \method{}, an inference-time method that replaces confidence-based unmasking with logic-role-guided unmasking. A lightweight classification head (4.2M parameters, 0.05\% of the base model) predicts the logical role of each masked position from the base model's hidden states with 98.4\% accuracy, and a dependency-ordered scheduler unmasks tokens in logical order. In zero-shot settings, \method{} improves LLaDA-8B-Instruct accuracy from 22.0\% to 60.7\% on GSM8K (+38.7\pp{}) and from 23.6\% to 29.2\% on MATH-500 (+5.6\pp{}), with less than 6\% speed overhead. However, with 8-shot chain-of-thought prompting, the baseline reaches ${\sim}$70\% and \method{} provides no additional improvement. Analysis reveals that few-shot prompting implicitly resolves the same ordering problem that \method{} explicitly addresses, and that fixed role-based ordering can cause premature commitment to numerical values before sufficient context is available. Our results characterize the Flexibility Trap as primarily a zero-shot phenomenon and identify context-adaptive ordering as a key direction for future work.
}
\vspace{8pt}

\section{Introduction}

The dominant paradigm in language modeling uses autoregressive~(AR) generation. A different approach has emerged with Masked Diffusion Language Models (MDLMs)~\citep{nie2025llada, shi2024mdlm, sahoo2024mdlm}, which generate text through iterative denoising from a fully masked sequence, offering parallel generation, bidirectional context, and the ability to revise tokens through remasking.

Despite these advantages, MDLMs cannot reason effectively. LLaDA-8B-Instruct~\citep{nie2025llada} achieves only ${\sim}22\%$ on GSM8K~\citep{cobbe2021gsm8k} in zero-shot settings, compared to ${>}70\%$ for AR models of similar size.

\citet{ni2026flexibility} identified the root cause as the \textbf{Flexibility Trap}: confidence-based unmasking systematically defers high-entropy logical connectives (``therefore,'' ``because,'' ``thus'') that serve as critical reasoning branching points. By filling easy tokens first, the model collapses the reasoning solution space before logical structure is established.

Existing fixes require expensive post-training: JustGRPO~\citep{ni2026flexibility} forces AR order during RL ($89.1\%$ GSM8K), d1~\citep{zhao2025d1} and SAPO~\citep{xie2025sapo} use reinforcement learning, ATPO~\citep{chen2025atpo} adaptively allocates gradient budget. All change model weights.

We propose \method{}, a different approach: rather than retraining the model, we fix the \emph{generation strategy} at inference time with three components:
\begin{enumerate}[leftmargin=2em, topsep=2pt, itemsep=1pt]
    \item A \textbf{Logic Role Classification Head}: a 2-layer MLP (4.2M params) classifying each masked position into five logical roles with 98.4\% accuracy.
    \item A \textbf{Dependency-Ordered Scheduler}: unmasks premises before connectives before derived steps before conclusions, preserving parallel generation within each role group.
    \item A \textbf{Priority Scoring Function}: weighted combination of role-based priority and confidence.
\end{enumerate}

In zero-shot settings, \method{} improves GSM8K from 22.0\% to 60.7\% (+38.7\pp{}) and MATH-500 from 23.6\% to 29.2\% (+5.6\pp{}). However, with 8-shot prompting the improvement vanishes, revealing that the Flexibility Trap is primarily a zero-shot phenomenon. We analyze why and identify directions for context-adaptive ordering.

\begin{figure}[h!]
\centering
\vspace{-4pt}
\includegraphics[width=0.95\textwidth]{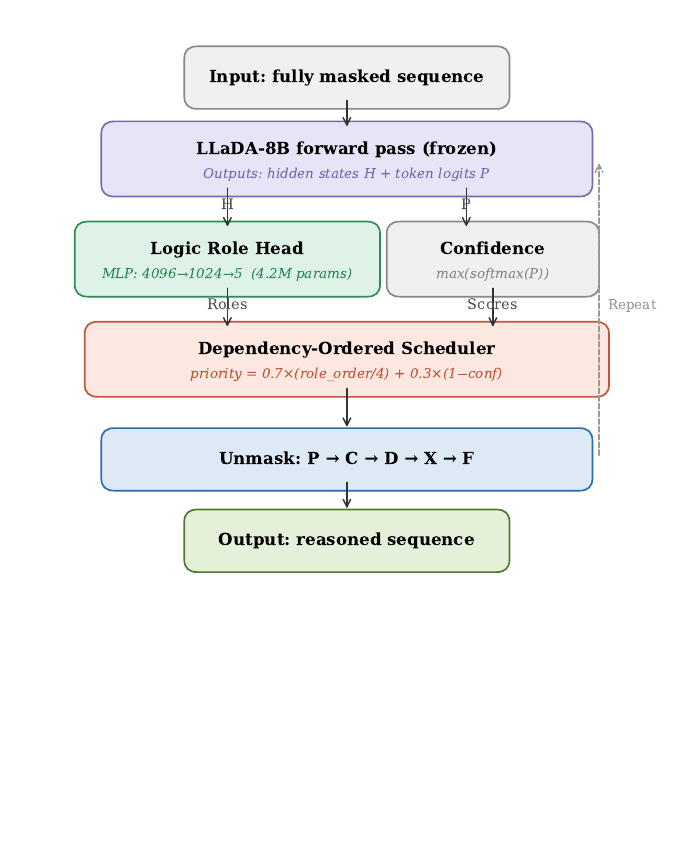}
\vspace{-4pt}
\caption{\method{} system architecture. The frozen LLaDA model produces hidden states and logits. The logic role head classifies each masked position. The dependency scheduler computes priority scores and unmasks tokens in logical dependency order.}
\label{fig:architecture}
\end{figure}
\FloatBarrier

\section{Related Work}

\textbf{Masked diffusion language models.}\quad LLaDA~\citep{nie2025llada} demonstrated competitive performance using Llama-3 with bidirectional attention. ReFusion~\citep{li2025refusion} introduced slot-level decoding. Dream~\citep{dream2025} and Block Diffusion~\citep{arriola2025block} explored hybrid architectures. A3~\citep{du2026a3} showed AR models can rival diffusion at any-order generation.

\textbf{Reasoning via RL.}\quad d1~\citep{zhao2025d1} achieved ${\sim}84.5\%$ GSM8K with diffu-GRPO. JustGRPO~\citep{ni2026flexibility} reached 89.1\% by treating MDLMs as AR during RL. SAPO~\citep{xie2025sapo} and ATPO~\citep{chen2025atpo} refined the approach. All modify model weights.

\textbf{Inference-time approaches.}\quad DoT~\citep{ye2024dot} distributes CoT across timesteps but requires training from scratch. DOS~\citep{zhou2026dos} uses attention as dependency proxies. \method{} differs by using an explicitly trained classifier predicting \emph{semantic logical roles}.

\textbf{Theory.}\quad \citet{feng2026theoretical} proved MDLMs need $\Omega(L)$ steps for low sequence error rate, suggesting generation \emph{order} matters more than step count.

\section{Method}

\subsection{Preliminaries}
In MDLMs, given prompt $\mathbf{q}$, the model initializes $\mathbf{x}^{(0)} = [\texttt{MASK}]^{L_g}$. At step $t$, it produces $\mathbf{H}^{(t)} \in \mathbb{R}^{L \times D}$ and $\mathbf{P}^{(t)} \in \mathbb{R}^{L \times |V|}$. Standard decoding unmasks the $K$ positions with highest $\mathrm{conf}(i) = \max_v P_{i,v}$.

\subsection{Logic Role Classification Head}
We introduce $f_\phi: \mathbb{R}^D \rightarrow \mathbb{R}^R$ predicting $R{=}5$ logical roles:
\vspace{2pt}
\begin{center}\small
\begin{tabular}{@{}clp{6.5cm}@{}}
\toprule
\textbf{ID} & \textbf{Role} & \textbf{Definition} \\
\midrule
0 & \premise{} & Given facts, known values, problem conditions \\
1 & \connective{} & Logical links: ``therefore,'' ``so,'' ``because'' \\
2 & \derived{} & Computed or inferred values \\
3 & \conclusion{} & Final answer or result \\
4 & \filler{} & Articles, punctuation, formatting \\
\bottomrule
\end{tabular}
\end{center}
\vspace{2pt}
Architecture (2-layer MLP with input LayerNorm):
\begin{equation}
f_\phi(\mathbf{h}) = W_2 \cdot \mathrm{Dropout}\!\big(\mathrm{GELU}(W_1 \cdot \mathrm{LayerNorm}(\mathbf{h}) + b_1)\big) + b_2
\end{equation}
$W_1 {\in} \mathbb{R}^{(D/4) \times D}$, $W_2 {\in} \mathbb{R}^{R \times (D/4)}$. For LLaDA-8B ($D{=}4096$, $R{=}5$): ${\sim}4.2$M params (0.05\% of base).

\textbf{Training data.}\quad Two-pass labeling of 7,473 GSM8K solutions (891,432 tokens): (1)~sentence-level role classification; (2)~token-level connective override. Distribution: \derived{} 93.6\%, \conclusion{} 3.9\%, \connective{} 1.3\%, \premise{} 0.8\%, \filler{} 0.4\%. Class-weighted CE with 10$\times$ for \connective{}.

\textbf{Training.}\quad Base model frozen. Random masking (${\sim}\mathcal{U}[0.3,0.9]$), forward pass, train $f_\phi$ on hidden states. 30 min on 1$\times$H100. \textbf{98.4\%} validation accuracy.

\subsection{Dependency-Ordered Scheduler}
Priority score for masked position $i$:
\begin{equation}
\mathrm{priority}(i) = w_r \cdot \frac{\mathrm{role\_order}(r_i)}{R - 1} + w_c \cdot \big(1 - \mathrm{conf}(i)\big)
\end{equation}
$\mathrm{role\_order}$: \premise{}${=}0$, \connective{}${=}1$, \derived{}${=}2$, \conclusion{}${=}3$, \filler{}${=}4$. We use $w_r{=}0.7$, $w_c{=}0.3$. Select $K = \lceil L_g / N \rceil$ lowest-priority positions per step.

\begin{figure}[h!]
\centering
\includegraphics[width=\textwidth]{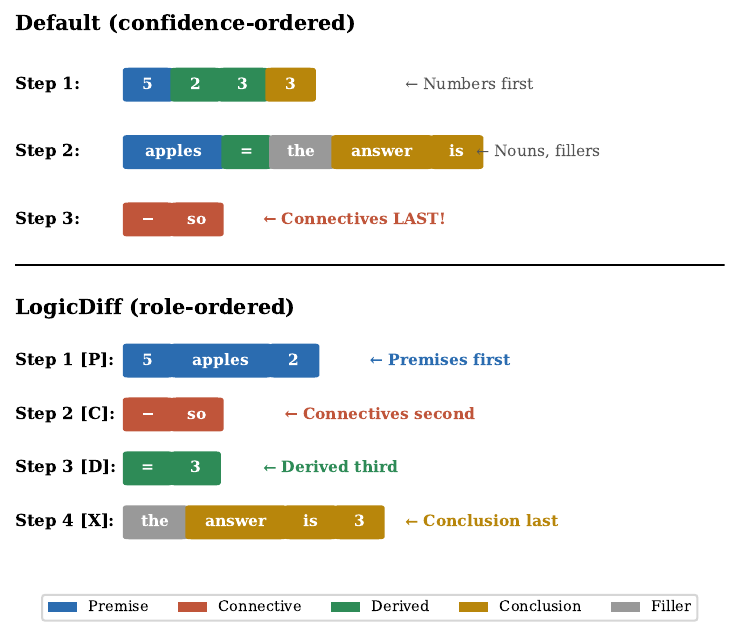}
\caption{Unmasking order comparison. \textbf{Top:} Default confidence-based unmasking generates numbers first and defers connectives to the last step. \textbf{Bottom:} \method{} unmasks premises first, then connectives, then derived results, then conclusions.}
\label{fig:comparison}
\end{figure}
\FloatBarrier

\subsection{Generation Algorithm}
Algorithm~\ref{alg:logicdiff} summarizes the complete generation procedure.
\begin{algorithm}[h!]
\caption{\method{} Generation}
\label{alg:logicdiff}
\begin{algorithmic}[1]
\REQUIRE Prompt $\mathbf{q}$, frozen model $M_\theta$, role head $f_\phi$, steps $N$, length $L_g$
\STATE $\mathbf{x} \leftarrow [\mathbf{q};\ \texttt{MASK}^{L_g}]$; $K \leftarrow \lceil L_g / N \rceil$
\FOR{$t = 1$ \TO $N$}
    \STATE $\mathbf{H}, \mathbf{P} \leftarrow M_\theta(\mathbf{x})$ \COMMENT{Forward pass (frozen)}
    \STATE $\mathcal{M} \leftarrow \{i : x_i = \texttt{MASK}\}$
    \IF{$|\mathcal{M}| = 0$} \STATE \textbf{break} \ENDIF
    \FOR{$i \in \mathcal{M}$}
        \STATE $r_i \leftarrow \arg\max f_\phi(\mathbf{h}_i)$; $s_i \leftarrow 0.7 \cdot \frac{\mathrm{role\_order}(r_i)}{4} + 0.3 \cdot (1 - \max_v P_{i,v})$
    \ENDFOR
    \STATE $\mathcal{U} \leftarrow \mathrm{bottom\text{-}K}(\{s_i\})$
    \FOR{$i \in \mathcal{U}$}
        \STATE $x_i \leftarrow \arg\max_v P_{i,v}$ \COMMENT{Unmask}
    \ENDFOR
\ENDFOR
\RETURN $\mathbf{x}$
\end{algorithmic}
\end{algorithm}
\FloatBarrier

\section{Experiments}

\subsection{Setup}
\textbf{Base model.}\quad LLaDA-8B-Instruct~\citep{nie2025llada}, 8B params, all frozen.
\textbf{Role head.}\quad Trained on 7,473 GSM8K solutions; same checkpoint for all experiments.
\textbf{Benchmarks.}\quad GSM8K~\citep{cobbe2021gsm8k} (1,319 problems), MATH-500~\citep{hendrycks2021math} (500 problems).

\subsection{Zero-Shot Results}

\begin{table}[h!]
\centering
\caption{\textbf{Zero-shot results} (0-shot, 256 max tokens, 256 steps). \method{} uses the same frozen model with no RL and no fine-tuning.}
\label{tab:zeroshot}
\small
\begin{tabular}{@{}lcccc@{}}
\toprule
\textbf{Method} & \textbf{GSM8K} & \textbf{MATH-500} & \textbf{Base Modified} & \textbf{Speed} \\
\midrule
LLaDA Baseline & 22.0\% (290/1319) & 23.6\% (118/500) & No & 0.18 ex/s \\
\method{} + Consistency & 3.0\% (3/100)$^\dagger$ & --- & No & 0.09 ex/s \\
\textbf{\method{}} & \textbf{60.7\% (800/1319)} & \textbf{29.2\% (146/500)} & \textbf{No} & \textbf{0.17 ex/s} \\
\midrule
& \textbf{+38.7\pp{}} & \textbf{+5.6\pp{}} & & ${<}6\%$ slower \\
\bottomrule
\multicolumn{5}{@{}l}{\footnotesize $^\dagger$ 100-example subset; disabled due to catastrophic failure.}
\end{tabular}
\end{table}

\begin{figure}[h!]
\centering
\includegraphics[width=0.85\textwidth]{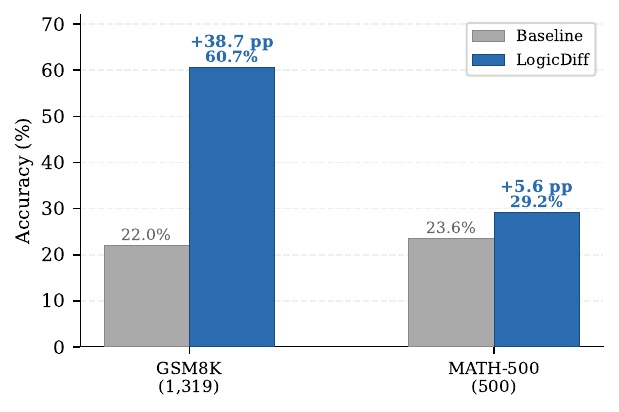}
\caption{Zero-shot accuracy on GSM8K and MATH-500.}
\label{fig:results}
\end{figure}
\FloatBarrier

In zero-shot settings, \method{} achieves \textbf{60.7\%} on GSM8K (+38.7\pp{} over baseline), solving 510 additional problems with ${<}6\%$ speed overhead. On MATH-500, +5.6\pp{} using the same role head without retraining.

\subsection{Effect of Few-Shot Prompting}

A critical question is whether \method{}'s improvement persists under standard evaluation conditions. We conducted experiments using 8-shot chain-of-thought prompting with 512 max tokens and 256 denoising steps, matching the official LLaDA evaluation protocol.

\begin{table}[h!]
\centering
\caption{\textbf{\method{} under different prompting conditions.} With 0-shot prompting, \method{} provides a large improvement. With 8-shot prompting, the baseline is already strong and \method{} provides no additional benefit.}
\label{tab:fewshot}
\small
\begin{tabular}{@{}llcc@{}}
\toprule
\textbf{Setting} & \textbf{Method} & \textbf{GSM8K} & \textbf{Tokens / Steps} \\
\midrule
0-shot & Baseline & 22.0\% & 256 / 256 \\
0-shot & \method{} & \textbf{60.7\%} (+38.7\pp{}) & 256 / 256 \\
\midrule
8-shot & Baseline & 70.0\% & 512 / 256 \\
8-shot & \method{} & 70.0\% (+0.0\pp{}) & 512 / 256 \\
\bottomrule
\end{tabular}
\end{table}
\FloatBarrier

The improvement vanishes entirely with 8-shot prompting. Further analysis on 30 problems that the 0-shot baseline answered incorrectly reveals that the 8-shot baseline solves 96.7\% of them (29/30), while \method{} solves 90.0\% (27/30)---a slight regression.

\subsection{Failure Analysis: Premature Numerical Commitment}

Examining problems where \method{} underperforms the 8-shot baseline reveals a consistent failure mode. \method{}'s role ordering forces numerical values to unmask early in the denoising process, but in early steps insufficient context has been revealed for the model to determine these values correctly.

For example, on a discount calculation problem, the baseline correctly generates ``75\% of the price'' by first establishing surrounding context through confidence-based ordering. \method{} forces the percentage to commit early, generating ``55\%'' instead---a plausible but incorrect value that propagates through the remaining computation. On a running speed problem, \method{} generates ``15 mph'' instead of ``1.5 mph'' for the same reason: the numerical value was committed before surrounding tokens provided disambiguating context.

This pattern recurs across all failure cases: a numerical value is committed too early, before the bidirectional context needed to determine it correctly has been established. The core assumption of \method{}---that premise tokens should be filled first---backfires when the model needs surrounding context to compute premise values correctly.

\subsection{Comparison with Existing Methods}

\begin{table}[h!]
\centering
\caption{\textbf{Method comparison} (0-shot setting). \method{} is the only inference-time method requiring no base model modification and no RL. Note that with 8-shot prompting, the LLaDA baseline reaches ${\sim}$70--78\% without any method modifications.}
\label{tab:comparison}
\small
\begin{tabular}{@{}lccccc@{}}
\toprule
\textbf{Method} & \textbf{GSM8K} & \textbf{MATH} & \textbf{RL} & \textbf{Modifies Base} & \textbf{Cost} \\
\midrule
LLaDA Baseline (0-shot) & 22.0\% & 23.6\% & No & No & 0 \\
LLaDA Baseline (8-shot) & $\sim$78\% & $\sim$27\% & No & No & 0 \\
DOS~\citep{zhou2026dos} & --- & --- & No & No & 0 \\
\midrule
d1~\citep{zhao2025d1} & $\sim$84.5\% & $\sim$41.0\% & Yes & Yes & Days (8$\times$A100) \\
JustGRPO~\citep{ni2026flexibility} & 89.1\% & 45.1\% & Yes & Yes & Days (8$\times$A100) \\
\midrule
\textbf{\method{} (0-shot)} & \textbf{60.7\%} & \textbf{29.2\%} & \textbf{No} & \textbf{No} & \textbf{30 min} \\
\bottomrule
\end{tabular}
\end{table}
\FloatBarrier

\subsection{Role Head Statistics}
Table~\ref{tab:rolehead} presents the training data distribution and classification performance.
\begin{table}[h!]
\centering
\caption{Training data and role head performance.}
\label{tab:rolehead}
\small
\begin{tabular}{@{}lrrr@{}}
\toprule
\textbf{Role} & \textbf{Count} & \textbf{Distribution} & \textbf{Weight} \\
\midrule
\premise{} & 7,512 & 0.8\% & 1.0 \\
\connective{} & 11,453 & 1.3\% & 10.0 \\
\derived{} & 834,290 & 93.6\% & 1.0 \\
\conclusion{} & 34,581 & 3.9\% & 2.0 \\
\filler{} & 3,596 & 0.4\% & 0.5 \\
\midrule
\textbf{Total} & \textbf{891,432} & & \\
\midrule
\multicolumn{3}{@{}l}{Validation accuracy} & \textbf{98.4\%} \\
\bottomrule
\end{tabular}
\end{table}
\FloatBarrier

\section{Analysis}

\subsection{Why Few-Shot Prompting Resolves the Flexibility Trap}

\method{} and few-shot prompting address the same underlying problem---establishing reasoning structure---through different mechanisms:

\begin{itemize}[leftmargin=2em, topsep=2pt, itemsep=2pt]
    \item \textbf{Few-shot prompting} provides explicit reasoning templates in the prompt. The model internalizes the logical ordering from examples and generates tokens in a reasoning-appropriate order even with confidence-based unmasking.
    \item \textbf{\method{}} enforces logical ordering externally through the sampler, compensating for the absence of reasoning templates. This is effective in zero-shot settings but redundant---and occasionally harmful---when templates are already provided.
\end{itemize}

This has two implications. First, the Flexibility Trap identified by \citet{ni2026flexibility} is primarily a zero-shot phenomenon: few-shot prompting implicitly resolves the ordering problem. Second, fixed role-based ordering is too rigid for settings with strong context---the model sometimes needs to build context before committing to numerical values, which confidence-based ordering naturally allows.

\subsection{Additional Analysis}

\textbf{GSM8K vs.\ MATH-500 gap.}\quad In zero-shot settings, GSM8K shows +38.7\pp{} while MATH-500 shows +5.6\pp{}. GSM8K follows a clear premise$\,{\to}\,$conclusion pattern that maps cleanly onto our taxonomy. MATH-500 involves complex algebraic manipulation where boundaries are less clear.

\textbf{DOS comparison.}\quad DOS~\citep{zhou2026dos} uses attention as statistical proxies (content-agnostic). \method{} uses a trained classifier predicting semantic roles (content-aware). Both face the same limitation: fixed ordering strategies cannot adapt to varying levels of available context.

\textbf{Consistency checker.}\quad CE-threshold remasking drops accuracy from 64\% to 3\%. The checker remasks correct but unusual tokens, creating a destructive cycle.

\section{Discussion}

\textbf{Implications.}\quad Our results paint a nuanced picture. In zero-shot settings, unmasking order has a dramatic effect on reasoning (+38.7\pp{}), confirming that MDLMs possess latent reasoning capabilities masked by suboptimal generation strategy. However, few-shot prompting---a simpler intervention---achieves even stronger results (${\sim}$70--78\%) by providing reasoning templates that implicitly resolve the ordering problem. This suggests that the Flexibility Trap, while real, may be less practically important than initially thought, since few-shot prompting is widely available in most deployment scenarios.

\textbf{Toward adaptive ordering.}\quad The key failure mode---premature numerical commitment---points to a concrete direction: \emph{context-adaptive ordering} that modulates the role-confidence weighting based on available context. In zero-shot settings, the scheduler should weight roles heavily; with few-shot context, it should defer more to confidence. Learning this balance end-to-end is a promising direction.

\textbf{Limitations.}\quad (1)~\method{}'s improvement is confined to zero-shot settings; with 8-shot prompting, it provides no benefit and can slightly hurt. (2)~The five-role taxonomy is coarse and trained only on GSM8K. (3)~Evaluated only on LLaDA-8B-Instruct. (4)~The consistency checker failed. (5)~The 0-shot evaluation uses 256 max tokens, which is shorter than the standard 512+ tokens used in official benchmarks.

\section{Conclusion}

We presented \method{}, an inference-time method that replaces confidence-based unmasking with logic-role-guided unmasking in masked diffusion language models. In zero-shot settings, \method{} improves GSM8K accuracy from 22.0\% to 60.7\% (+38.7\pp{}) on frozen LLaDA-8B-Instruct, demonstrating that the Flexibility Trap can be addressed without RL or weight modification. However, with 8-shot prompting, the baseline reaches ${\sim}$70\% and \method{} provides no additional gain, revealing that few-shot prompting implicitly resolves the same ordering problem. Analysis of failure cases identifies premature numerical commitment as the key limitation of fixed role ordering. Our results characterize the Flexibility Trap as primarily a zero-shot phenomenon, and suggest that context-adaptive unmasking strategies---rather than fixed orderings---are needed to improve MDLM reasoning across all settings.

{\small
\bibliographystyle{plainnat}

}

\end{document}